# Comparison of Neural Network based Soft Computing Techniques for Electromagnetic Modeling of a Microstrip Patch Antenna

Yuvraj Singh Malhi[1] and Navneet Gupta[2]

Department of Electrical and Electronics, Birla Institute of Technology and Science Pilani, Rajasthan, India, 333031

```
yuvrajmalhi30@gmail.com[1]
ngupta@pilani.bits-pilani.ac.in[2]
```

**Abstract.** This paper presents the comparison of various neural networks and algorithms based on accuracy, quickness, and consistency for antenna modelling. Using MATLAB's Nntool, 22 different combinations of networks and training algorithms are used to predict the dimensions of a rectangular microstrip antenna using dielectric constant, height of substrate, and frequency of operation as input. Comparison and characterization of networks is done based on accuracy, mean square error, and training time. Algorithms, on the other hand, are analyzed by their accuracy, speed, reliability, and smoothness in the training process. Finally, these results are analyzed, and recommendations are made for each neural network and algorithm based on uses, advantages, and disadvantages. For example, it is observed that Reduced Radial Bias network is the most accurate network and Scaled Conjugate Gradient is the most reliable algorithm for electromagnetic modelling. This paper will help a researcher find the optimum network and algorithm directly without doing time-taking experimentation.

## 1  Introduction

Microstrip antennas are among the most common antennas in use because they offer advantages like easy installation, variable size, low cost, easy modelling, wide range of resonant frequencies, and conformable shape. Microstrip antennas can be categorized based on patch shape (rectangular, elliptical, circular ring, or dipole) or based on feed methods (probe, microstrip line, aperture coupled, or proximity control) [1]. In this paper, a rectangular patch antenna with Microstrip feed has been chosen because of the wide acceptability of its transmission line model analysis.

   Transmission line model gives relatively complex mathematical formulas for finding the physical parameters (length and width) based on some design parameters (frequency of operation and height and dielectric constant of substrate). The calculations of transmission line model are computationally heavy, and this is where the role of soft computing comes in, particularly with Artificial Neural Networks (ANN). ANN is an artificial system that consists of a structure with similar operation as that of the human brain [2]. It consists of many relatively simple nonlinear functional blocks (neurons), which process some input signals into an output signal using an activation function. All existing Using data with known results, training algorithms are used to train the ANN for predicting unknown results. Therefore, ANNs can be classified based on network structure and training algorithms. The motivation behind this paper is to bridge the gap between researchers and the available tools in ANNs that may help improve performance of models and designs.

## 2   Literature Review and Motivation

Though some researchers have done electromagnetic modelling of rectangular patch microstrip antennas using ANNs [4-6], none have quantitatively compared the performance of their models using different neural networks and algorithms. In some cases, comparison between few algorithms is considered but the algorithms were chosen randomly and not all algorithms were explored [13]. It is also observed that there are more applications of radial basis and feed forward networks but hardly any article explains the reasons for choosing these and not other available networks. This may be because many, if not all, researchers blindly choose an algorithm which gives a relatively low error without considering the other alternatives that can simultaneously reduce the training time and error by over 75%. In some cases, more difficult soft computing techniques for antenna design have been explored like the use of fuzzy logic which yields lower accuracy and is difficult to implement [14]. Sometimes ANNs are not even considered [15] and PSO is used despite its bad performance and implementation difficulty compared to ANN [16].

One reason why ANNs are not used is because their parameter tuning is difficult, time-taking, and may require mathematical expertise. Even when ANN networks and algorithms are used, they are not always compared quantitatively before finding the best network and algorithm. This is because such comparison takes time, effort, and mathematical knowledge of hyper-parameters. Another reason why ANNs are not properly explored is because there are simply too many algorithms in use, and identifying the optimum one is very tedious. For electromagnetic modelling itself more than 10 different algorithms have been used [3]. These are just some of the reasons preventing designs from achieving their full potential.

From this paper, it is expected that researchers will simply look at the analysis of each network and algorithm in terms of their speed, accuracy, smoothness, and reliability and then easily identify the most suitable combination for their application. Overall, the process of finding the optimal network and algorithm will be shortened greatly.

## 3   Network Model and Data Generation

For the comparison of different networks and algorithms, this paper applies ANN in the use case of predicting width and length of a rectangular microstrip antenna by using frequency of operation, dielectric constant of substrate, and height of substrate as input parameters. The comparison is done between 22 combinations of networks and algorithms. To see practical performance, after training each of the 22 combinations, the resulting ANN is tested for 5 different real-world applications (see Table 3). To ensure reliability of results, each trial is performed 5 times and average values of mean squared error (MSE) and training time is calculated.

MATLAB's Nntool is chosen as the platform for this analysis because Nntool, unlike python or R, does extremely easy construction of neural networks with a smaller training data and automatically generates MSE graphs.

The transmission line model analysis is widely accepted for mathematical characterization of rectangular microstrip antennas due to its clear physical insight and its ability to effectively approximate antenna parameters like length and dielectric [6-7]. Figure 1 shows the schematic view of a rectangular microstrip patch antenna. The physical and design properties in this model are determined by Eq. 1-5, as taken from [3].

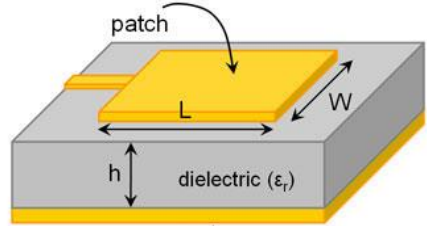

**Fig. 1.** Rectangular microstrip patch antenna.

Width of patch:
$$W = \frac{c}{2f_r}\sqrt{\frac{2}{\varepsilon_r + 1}} \quad (1)$$

Effective Dielectric constant:
$$\varepsilon_{eff} = \frac{\varepsilon_r + 1}{2} + \frac{\varepsilon_r - 1}{2}\left(\frac{1}{\sqrt{1 + \frac{2h}{W}}}\right) \quad (2)$$

Effective length:
$$L_{eff} = \frac{C}{2f_r\sqrt{\varepsilon_{eff}}} \quad (3)$$

Change in Length:
$$\Delta L = h * 0.412 * \frac{(\varepsilon_{eff} + 0.3)\left(\frac{W}{h} + 0.264\right)}{(\varepsilon_{eff} - 0.258)\left(\frac{W}{h} + 0.8\right)} \quad (4)$$

Length of patch:
$$L = L_{eff} - 2\Delta L \quad (5)$$

Here, $c$ is speed of light, $f_r$ is center frequency, and $h$ and $\varepsilon_r$ are the height and dielectric constant of the substrate, respectively. The neural network model of microstrip antennas is illustrated in Fig. 2 where there are 3 input variables: $h$, $f_r$, and $\varepsilon_r$, and predicts 2 output variables $W$ (width of patch) and $L$ (length of patch) of the patch. Out of the 3 inputs, the height of substrate ($h$) is fixed at the 1.5 mm since this is the standard size of all manufacturers.

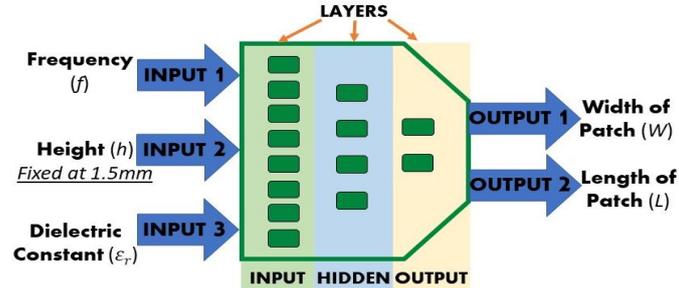

**Fig. 2.** Network model.

For a neural network to model only 3 inputs, a lot of neurons are not required in every layer. However, a hidden layer is required for modelling the complex mathematical equations. So, wherever possible, the size of the network is kept as *8 × 4 × 2* where 8 neurons are used after the input layer, 4 are used at the hidden layer, and 2 are used in the hidden layer. Figure 2 depicts this antenna but the connections of neurons in have not been shown Fig. 2 as they may differ for different types of networks.

Range of microstrip antenna frequencies for some common applications [9] and some common dielectric constants [10,11] are stated in Table 1.

**Table 1.** Applications and dielectric constants.

| Applications and frequencies (GHz) | | Substrates and Di-electric constants | |
|---|---|---|---|
| Application | Frequency | Substrate | Dielectric constant |
| Wi-Fi – 802.11n | 2.5 | RT Duroid | 2.2 |
| Wi-Fi – 802.11ac | 5 | | |
| Wi-Fi – 802.11ad | 5 and 60 | Tacon TLX 6 | 2.65 |
| Bluetooth | 2.48 | | |
| 2G, 3G | 0.9 and 1.8 | Roger 4350 | 2.65 |
| 4G | 0.85 and 2.3 | | |
| WiMAX | 2.3 | FR4 Glass Epoxy | 4.36 |
| Satellite | 12 to 18 ( ~15) | | |
| GPS – L1 | 1.51 | Duroid 6010 | 10.5 |
| GPS – L5 | 1.17 | | |

To train the neural network, training data needs to be generated. Since Nntool re-generates more random samples within the training set, a sample size above 1000 would be redundant and will slow down the training process. Operating frequency needs to be precise for every application, so the frequency is varied in very small steps of 0.0975 GHz beginning from 0.5 GHz and up to 20 GHz which gives 200 variations. The dielectric constant is varied from 1.2 to 11 in steps of 0.196 which gives 50 variations. Thus, the total number of input training samples become 200× 50 = 1000 training samples (see Fig. 3) and or each sample, Eq. 1-5 are used to find the target width and length.

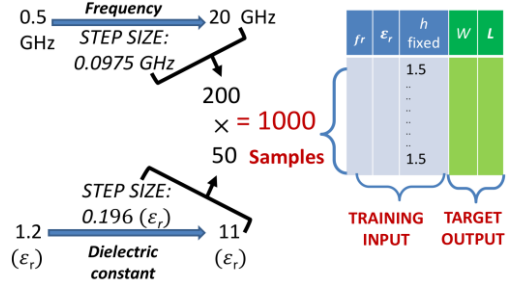

**Fig. 3.** Training set.

## 4   Networks and Algorithms Selection

Out of the 40 algorithms initially considered, 18 were eliminated because of inaccuracy or inconsistency (tendency to get stuck in local minima). But despite low accuracy, generalized regression is kept because it is extremely fast and adds variety to the mix. Table 2 mentions the different training functions and networks used. Gradient descent algorithms were later excluded because of poor performance in every scenario.

**Table 2.** Training functions and networks

| List of training functions used | List of network types used |
|---|---|
| • Levenberg Marquardt (LM)<br>• Resilient Backpropagation (RP)<br>• Scaled Conjugate Gradient (SCG)<br>• Conjugate Gradient with Powell/Beale (CGB)<br>• Fletcher-Powell Conjugate (CGF)<br>• Polak-Ribiére Conjugate Gradient (CGP)<br>• Gradient Descent Algorithms (GD, GDX, GDM) | • Cascade Forward Backprop. (CF)<br>• Elman Backpropagation (EL)<br>• Feed Forward Backpropagation (FF)<br>• Generalised Regression (GR)<br>• Layer recurrent (LR)<br>• Radial Bias (Reduced) (RBR) |

A function calculates the accuracy of each sample according to a custom formula in Eq. 6. MSE for the sample is calculated by Nntool using the conventional formula in Eq. 7. Here, $Y_o$ is the mathematically computed target, $Y_p$ is the predicted output value, and N is the number of samples.

$$\text{Mean Squared Error (MSE)} = \sum_{n=1}^{N} \frac{(Y_o - Y_p)^2}{N} \tag{6}$$

$$\text{Accuracy (\%)} = \frac{|Y_o - |Y_o - Y_p||}{Y_o} \times 100 \tag{7}$$

Out of all applications and substrates in Table 1, only 5 combinations were chosen as use cases to demonstrate the performance of the different ANNs. The selected applications, substrates, and their target values of width and length are specified in Table 3.

**Table 3.** Target values for five chosen applications.

| RT Duroid for Wi-Fi 802.11ac | | TLX 6 for Bluetooth | | Roger 4350 for GPS L5 | | FR 4 for WiMAX | | Duroid 6010 for 3G | |
|---|---|---|---|---|---|---|---|---|---|
| W (mm) | L (mm) | W (mm) | L (mm) | W (mm) | L (mm) | W (mm) | L (mm) | W (mm) | L (mm) |
| 23.717 | 19.398 | 44.772 | 36.587 | 85.660 | 68.429 | 39.837 | 30.934 | 34.752 | 25.622 |

## 5 Simulation Results and Discussion

Table 4 shows the predicted values of W, L, training time, and average accuracy for each combination of network and algorithm averaged over 5 trials. ATT stands for average training time and AA stands for average accuracy.

**Table 4.** Simulation performance values.

| Network + Training Algo. | M.S.E. (mm) | ATT (s) | RTD for Wi-Fi 802.11ac | | TLX 6 for Bluetooth | | Roger 4350 for GPS L5 | | FR 4 for WiMAX | | Duroid 6010 for 3G | | AA (%) |
|---|---|---|---|---|---|---|---|---|---|---|---|---|---|
| | | | $W_{avg}$ (mm) | $L_{avg}$ (mm) | $W_{avg}$ (mm) | $L_{avg}$ (mm) | $W_{avg}$ (mm) | $L_{avg}$ (mm) | $W_{avg}$ (mm) | $L_{avg}$ (mm) | $W_{avg}$ (mm) | $L_{avg}$ (mm) | |
| *CF + LM* | 0.03 | 5 | 23.69 | 19.43 | 44.82 | 36.50 | 85.62 | 68.37 | 40.00 | 30.79 | 34.69 | 25.63 | *99.81* |
| *CF + RP* | 0.50 | 25 | 23.20 | 19.12 | 46.31 | 36.23 | 84.22 | 65.61 | 41.42 | 31.27 | 32.82 | 25.09 | *97.35* |
| *CF + SCG* | 0.04 | 60 | 23.74 | 19.36 | 44.91 | 36.22 | 85.57 | 68.21 | 40.01 | 30.95 | 34.88 | 25.73 | *99.69* |
| *CF + CGF* | 0.24 | 48 | 23.95 | 19.47 | 43.91 | 35.78 | 86.84 | 68.83 | 39.46 | 30.40 | 34.67 | 25.71 | *98.92* |
| *CF + CGP* | 0.1 | 17 | 23.97 | 19.63 | 44.79 | 36.60 | 86.17 | 68.71 | 39.57 | 30.77 | 34.89 | 26.13 | *99.28* |
| *EL + LM* | 0.02 | 75 | 23.79 | 19.47 | 44.65 | 36.66 | 85.80 | 68.49 | 39.76 | 30.96 | 34.79 | 25.50 | *99.90* |
| *EL + RP* | 0.43 | 35 | 23.96 | 19.64 | 46.50 | 38.21 | 82.13 | 67.05 | 40.90 | 31.35 | 32.77 | 25.11 | *98.81* |
| *EL + SCG* | 0.12 | 30 | 23.83 | 19.51 | 44.99 | 36.07 | 86.37 | 68.63 | 39.57 | 30.30 | 34.73 | 25.55 | *99.28* |
| *EL + CGP* | 1.26 | 10 | 24.04 | 18.33 | 46.52 | 37.55 | 83.49 | 68.09 | 39.97 | 31.80 | 33.09 | 25.63 | *97.56* |
| *EL + OSS* | 0.23 | $10^2$ | 24.05 | 19.42 | 44.91 | 36.45 | 86.05 | 68.70 | 39.58 | 30.72 | 34.59 | 25.66 | *99.49* |
| *GR* | 5.17 | $10^{-3}$ | 24.28 | 20.00 | 51.76 | 42.90 | 85.82 | 69.19 | 46.61 | 36.43 | 40.96 | 30.27 | *88.94* |
| *FF + LM* | 0.05 | 70 | 23.71 | 19.43 | 44.71 | 36.62 | 85.70 | 68.41 | 39.77 | 30.87 | 34.75 | 25.68 | *99.84* |
| *FF + RP* | 0.3 | 60 | 22.93 | 18.99 | 45.34 | 37.49 | 83.74 | 67.12 | 40.39 | 31.54 | 35.39 | 26.11 | *97.95* |
| *FF + SCG* | 0.03 | 68 | 23.63 | 19.40 | 44.91 | 36.36 | 86.58 | 68.94 | 39.73 | 30.46 | 34.51 | 25.67 | *99.42* |
| *FF + CGB* | 0.09 | 30 | 23.59 | 19.55 | 44.59 | 36.33 | 86.76 | 68.80 | 39.47 | 30.67 | 34.31 | 25.48 | *99.22* |
| *FF + CGP* | 0.13 | 20 | 23.77 | 19.83 | 44.51 | 36.52 | 85.31 | 68.08 | 39.76 | 30.49 | 34.54 | 25.54 | *99.33* |
| *LR + RP* | 0.08 | 33 | 23.78 | 19.64 | 45.45 | 37.00 | 85.73 | 69.01 | 39.96 | 30.60 | 34.01 | 25.81 | *99.05* |
| *LR + SCG* | 0.03 | $10^2$ | 23.69 | 19.40 | 44.73 | 36.73 | 86.03 | 68.51 | 39.71 | 31.03 | 34.70 | 25.62 | *99.80* |
| *LR + CGF* | 0.10 | 15 | 24.10 | 19.48 | 44.33 | 36.25 | 86.06 | 68.61 | 39.50 | 30.43 | 34.26 | 25.53 | *99.11* |
| *LR + CGP* | 0.24 | 30 | 24.10 | 20.03 | 44.34 | 36.98 | 86.06 | 69.73 | 40.23 | 31.22 | 35.34 | 25.63 | *99.32* |
| *0.001 RBR* | 0.001 | 300 | 23.74 | 19.42 | 44.76 | 36.58 | 85.60 | 68.38 | 39.83 | 30.93 | 34.81 | 25.66 | *99.927* |
| *0.0006 RBR* | 0.0006 | 1000 | 23.72 | 19.40 | 44.76 | 36.58 | 85.63 | 68.40 | 39.83 | 30.93 | 34.83 | 25.68 | *99.938* |

The training simulation graphs of 1 among the 5 training repetitions for each network and algorithm are shown in Fig. 4-9.

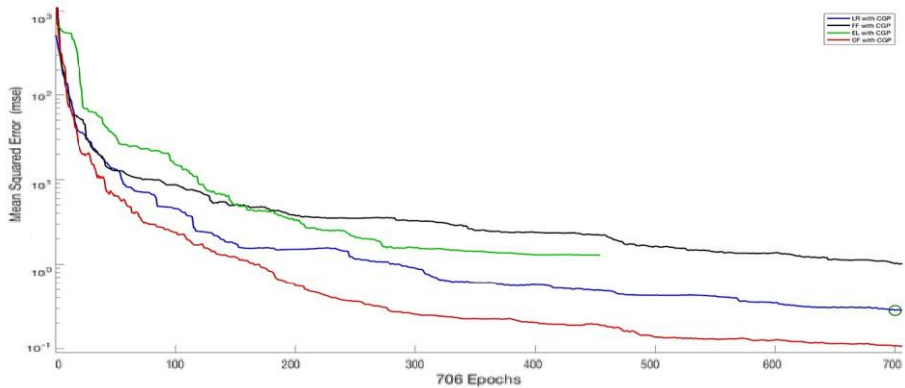

**Fig 4.** Variation of mean squared error (MSE) versus the number of epochs for CGP

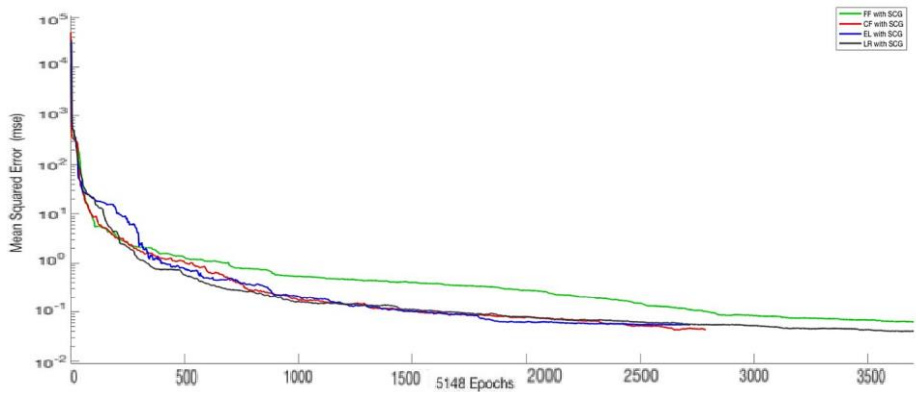

**Fig 5.** Variation of mean squared error (MSE) versus the number of epochs SCG.

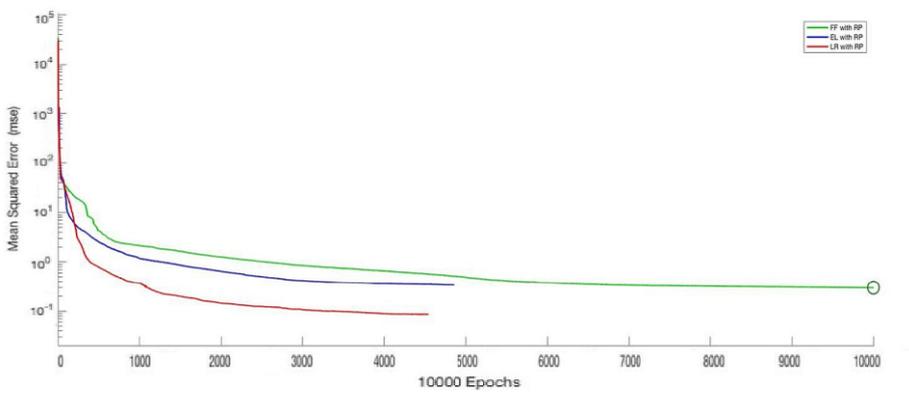

**Fig 6.** Variation of mean squared error (MSE) versus the number of epochs for RP.

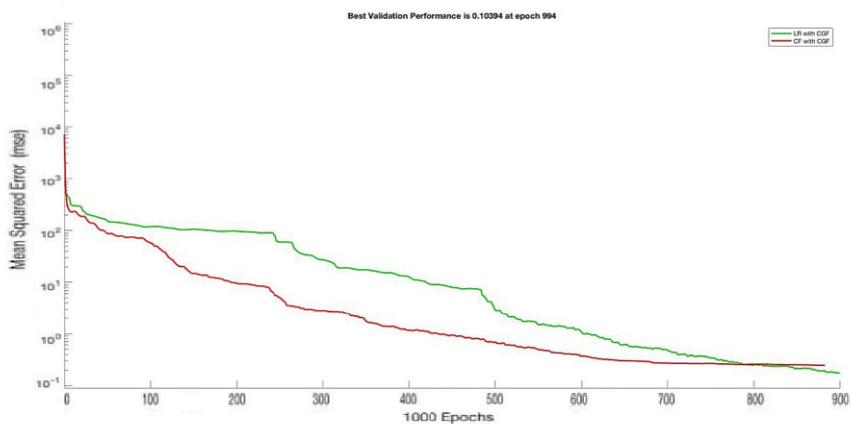

**Fig 7.** Variation of mean squared error (MSE) versus the number of epochs for CGF.

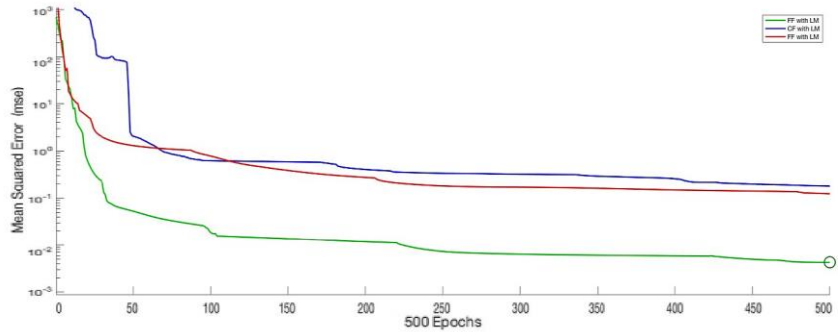

**Fig 8.** Variation of mean squared error (MSE) versus the number of epochs for LM.

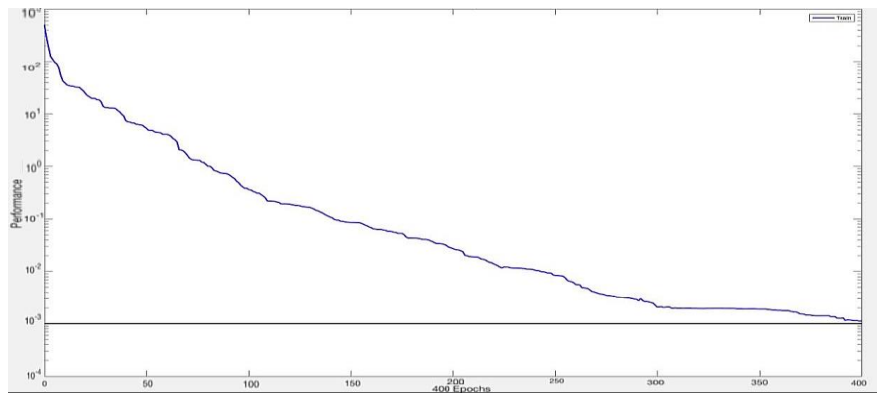

**Fig 9.** Variation of performance versus the number of epochs for RBR.

From Fig. 4-9 and table 4, following recommendations are made for networks and algorithms:

### 5.1 Discussion on Network Recommendations

Radial Basis Neural Network easily give accuracy of more than 99.9% and MSE less than 0.001. Also, RBR does not require any user input for the network size/ hyperparameters. In fact, RBR can even learn complex patterns without tuning, while still cutting the error to a tenth. However, RBR takes much more training time than others.

Generalized regression is the fastest and requires no extra input from user. But it gives an average accuracy of about 90% and MSE of 5.7. So, GR takes a 0.01% of the other training times and requires no parameter tuning but gives 10 times higher error.

CF (Cascade Forward Backpropagation) and LR (Layer recurrent) are similar performance networks with CF being a little faster and LR being smoother and more consistent. They both give decent accuracy levels (99%) and take relatively less time to converge. So, CF and LR are user friendly networks which have the flexibility to work with default parameters and always converge.

EL (Elman Backprop.) is consistent and works with all algorithms but is slow and prematurely terminates within 500 epochs, even with minimum gradient of $10^{-20}$.

FF (Feed Forward Backprop) is not recommended because of unreliability. FF takes many iterations to converge, gives large error and works only with CGF.

### 5.2 Discussion on Algorithm Recommendations

A good balance between time and error is given by SCG (Scaled Conjugate Gradient) algorithm which gives accuracy levels of more than 99%, takes 60s to converge and works will all networks. So, SCG is a consistent, accurate, and time savvy training algorithm which works with all networks but needs proper tuning of hyper-parameters like learning rate and minimum gradient.

CGP (Polak-Ribiére Conjugate Gradient) is a very fast algorithm, and its accuracy levels are greatly influenced by the network. It is prone to get stuck in local minima if the network or initialization of neuron weights is random. So, CGP is useful when one has the time to train multiple times with different networks.

RP (Resilient Backpropagation) is a quick algorithm but requires hyper-parameter tuning. Since RP's learning rate is large, it is very smooth and consistent but for the same reason, RP does not give good accuracy as exact minima is never reached.

CGB (Conjugate Gradient with Powell/Beale Restarts) is an accurate but extremely inconsistent algorithm which only works with FF and is therefore not recommended.

## 6 Conclusion and Future Work

In this paper the implementation and analysis of various ANN based soft computing techniques is presented. This paper helps in selecting and characterizing different networks and training algorithms for electromagnetic modelling. A robust qualitative experiment is done and recommendations for the use of different combinations of networks and algorithms are presented based on to speed, accuracy, and consistency. Overall, this paper bridges the gap between researchers and the powerful tools of neural networks.

Researchers in other domains may still have to treat ANNs as black boxes where the only way of choosing the best ANN is through time consuming experimentation. However, if such studies can be carried out in other domains, the selection process will be greatly optimized in other domains as well.

## 7 References

1. Balanis, C. A.: Antenna theory: Analysis and design (3rd ed.). John Wiley, NJ USA (2005).
2. Zhang, Z.: A gentle introduction to artificial neural networks. Annals of Translational Medicine (4), 370-370 (2016).
3. Choudhury, B., Jha, R.: Soft Computing in Electromagnetics: Methods and Applications. Cambridge University Press, Cambridge (2016).
4. Soliman, E.A., Bakr, M.H., Nikolova, N.K.: Modeling of microstrip lines using neural networks: Applications to the design and analysis of distributed microstrip circuits. Int J RF and Microwave Comp Aid Eng (14), 166-173 (2004).
5. Kushwah, V.S., Tomar, G.S.: Design and Analysis of Microstrip Patch Antennas Using Artificial Neural Network (2017).
6. Singh, A.K., Pandey, M.S.: Design and Analysis of Microstrip Patch Antennas Using Soft Computing, International Journal of Latest Trends in Engineering and Technology (2016).

7. Guney, K., Gultekin, S.S.: Artificial Neural Networks for Resonant Frequency Calculation of Rectangular Microstrip Antennas with Thin and Thick Substrates. International Journal of Infrared and Millimeter Waves (25), 1383–1399 (2004).
8. Werfelli, H., Tayari, K., Chaoui, M., Lahiani, M., Ghariani, H.: Design of rectangular microstrip patch antenna. 2nd International Conference on Advanced Technologies for Signal and Image Processing (ATSIP), 798-803 (2016).
9. Patel, S., & Shah, V., Kansara, M.: Comparative Study of 2G, 3G and 4G. International Journal of Scientific Research in Computer Science, (3) 2456-3307 (2018).
10. Jain, K., Gupta, K.: Different Substrates Use in Microstrip Patch Antenna-A Survey. International Journal of Science and Research, 2319-7064 (2014).
11. Khan, Anzar & Nema, Rajesh. Analysis of Five Different Dielectric Substrates on Microstrip Patch Antenna. International Journal of Computer Applications (2012).
12. Raida, Z. Modeling EM structures in neural network toolbox of MATLAB (2002).
13. Singh, B.K.: Design of rectangular microstrip patch antenna based on Artificial Neural Network algorithm, International Conference on Signal Processing and Networks (2), 6-9 (2015)
14. Mohammed H., Hamdi M.M., Rashid S.A., Shantaf A.M.: An Optimum Design of Square Microstrip Patch Antenna Based on Fuzzy Logic Rules. 2020 International Congress on Human-Computer Interaction, Optimization and Robotic Applications (2020).
15. Alhamadani, N.B., Abdelwahid, M.M.: Implementation of Microstrip Patch Antenna Using MATLAB. Journal of Applied Machines Electrical Electronics Computer Science and Communication Systems 2, no. 1 (March 29, 2021): 29 - 35. Accessed May 9 (2021).
16. Girija, H. S., Sudhakar, R., Kadhar K. M., Priya, T. S., Anand, G.: PSO Based Microstrip Patch Antenna Design for ISM Band. 6th International Conference on Advanced Computing and Communication Systems, 1209-1214 (2020).